\def\year{2022}\relax
\documentclass[letterpaper]{article} 
\usepackage{aaai22}  
\usepackage{times}  
\usepackage{helvet}  
\usepackage{courier}  
\usepackage[hyphens]{url}  
\usepackage{bbold}
\usepackage{dirtytalk}
\usepackage{graphicx} 
\usepackage{natbib}  
\usepackage{caption} 
\DeclareCaptionStyle{ruled}{labelfont=normalfont,labelsep=colon,strut=off} 
\usepackage{algorithm}
\usepackage{algorithmic}
\usepackage{tablefootnote}

\usepackage{newfloat}
\usepackage{listings}
\floatstyle{ruled}
\newfloat{listing}{tb}{lst}{}
\floatname{listing}{Listing}
\title{A Review of Uncertainty for Deep Reinforcement Learning}

\author {
  Owen Lockwood\textsuperscript{\rm 1}
  ,
   Mei Si \textsuperscript{\rm 2}
}
\affiliations {
\textsuperscript{\rm 1} Department of Computer Science, Rensselaer Polytechnic Institute\\
\textsuperscript{\rm 2} Department of Cognitive Science, Rensselaer Polytechnic Institute\\
    lockwo@rpi.edu, sim@rpi.edu
}

\begin{document}

\maketitle

\begin{abstract}

Uncertainty is ubiquitous in games, both in the agents playing games and often in the games themselves. Working with uncertainty is therefore an important component of successful deep reinforcement learning agents. While there has been substantial effort and progress in understanding and working with uncertainty for supervised learning, the body of literature for uncertainty aware deep reinforcement learning is less developed. While many of the same problems regarding uncertainty in neural networks for supervised learning remain for reinforcement learning, there are additional sources of uncertainty due to the nature of an interactable environment. In this work, we provide an overview motivating and presenting existing techniques in uncertainty aware deep reinforcement learning. These works show empirical benefits on a variety of reinforcement learning tasks. This work serves to help to centralize the disparate results and promote future research in this area. 

\end{abstract}

\section{Introduction}

Deep reinforcement learning techniques have become the state of the art methods for a number of games including games such as Chess, Go, Shogi \cite{silver2018general}, and
Dota 2 \cite{berner2019dota}
to name a few. However, the much of the standard suite of deep reinforcement learning algorithms have little to no awareness or quantification of the uncertainty in either the agent or the environment. This can lead to brittleness in cases in which the reinforcement learning agents perform unsuccessfully when encountering new situations. 
Building uncertainty aware agents is important for building robust and versatile agents.

The task of quantifying and incorporating uncertainty in neural networks is essential for a variety of tasks, such as imperfect information games (e.g. poker, SCII), increasing exploration of RL agents, flagging high uncertainty samples for human review (in real world deployments), guarding against adversarial examples,  active learning, etc. The techniques for supervised learning have substantially grown with the explosion of deep learning in the last decade \citep{gawlikowski2021survey, pearce2020uncertainty, kendall2017uncertainties, gal2016dropout}. 
These works usually focus on one of two areas of uncertainty, the uncertainty in the dataset, i.e. noise in the labels $y_{label} = y_{true} + \epsilon$, or the uncertainty of the neural network model. With the success of uncertainty modelling in the supervised learning field, reinforcement learning has begun to adopt some of these techniques. Building upon these approaches, uncertainty aware reinforcement learning has been able to so improvements in empirical performance and data efficiency. 

In this work we outline some of the advances being made in uncertainty aware deep reinforcement learning. These techniques have resulted in improved performance on hard exploration games, continuous control tasks, and hold the current state of the art performance for offline reinforcement learning.
We begin by outlining some of the background on deep reinforcement learning. We then connect uncertainty to deep reinforcement learning and how recent works have attempted to address uncertainty and the results that these works have shown.

\section{Reinforcement Learning}

\subsection{Markov Decision Process}

Reinforcement learning is often formalized via the Markov Decision Process (MDP), defined by the tuple $\langle S, A, R, P \rangle$. In this representation, $S$ is the set of states, $A$ the set of actions, $R$ is a function mapping states and actions to rewards, and $P$ is the transition probability (i.e. the probably that a given state follows another given some input action). $\gamma$ is sometimes included in this tuple and is used to denote the discount factor on the reward. 
The goal of reinforcement learning is to maximize the numerical reward signal \citep{sutton2018reinforcement}. In this framework, we write this objective function as $J(\pi) = \max_\pi \mathbb E_{a \sim \pi} \left [ \sum^T_i \gamma^i R(s_i, a_i) \right ]$. In other words, we wish to find a policy, $\pi$, such that it maximizes the sum of the discounted rewards. This can also be formulated as a maximization of the Q function. The Q function, also called the action value function \citep{watkins1992q}
, is defined as the expected return of a policy from a given state action pair.

\subsection{Model Free Algorithms}

In order to maximize this numerical reward, contemporary deep learning methods use a variety of techniques. Here we focus on model free algorithms. These algorithms generally fall two main categories: value based and actor-critic methods. The most common value based method is Deep Q Learning \citep{mnih2013playing}. Deep Q Learning replaces the dynamic programming approach to Q learning with a Deep Q Network (DQN) that learns to approximate the Q values of the actions given an input state. 
These Deep Q Networks are updated via the Mean-Squared Bellman Error, $\mathcal{L}(\theta) = \left ( r + \gamma \max Q(s_{i+1}, a_{i+1}, \theta) - Q(s_i, a_i, \theta) \right )^2$. 
This approach has a number  challenges, such as a moving target, over estimation of Q values, and wall clock time inefficiency. Numerous modifications have been proposed, with varying degrees of improvements \citep{hessel2018rainbow}. 
Distributional Q algorithms, such as QR-DQN \cite{dabney2018distributional}, predict the distribution of expected rewards $Z$, rather than the exact Q value. Note that this distribution is a poor estimation of the uncertainty \cite{osband2018randomized}.
However, even with these improvements limitations remain. The learning efficiency is generally lower than actor-critic methods and they are limited to discrete action environments.

Actor critic methods use both a policy neural network and a critic neural network to help inform the training of the policy. 
Deep Deterministic Policy Gradient (DDPG) \citep{lillicrap2015continuous} uses a deterministic policy network updated via information from the Deep Q Network by the deterministic policy gradient theorem, $\nabla J(\mu_\theta) = \mathbb E \left [ \nabla \mu_\theta (s) \nabla Q(s, \mu_\theta (s)) \right ] $. It relies on the Mean Squared Bellman Error to update the Q network.  
Stochastic policy algorithms parameterized a distribution over the action space, in opposition to the above deterministic policy. This distribution is traditionally represented as a Gaussian, with mean $\mu_\theta$ and standard deviation $\sigma_\theta$. 
Proximal Policy Optimization (PPO) is one such algorithm \citep{schulman2017proximal}. PPO relies on an advantage based critic, where $A(s, a) = Q(s, a) - V(s)$, implemented via a value neural network which allows an approximation of the advantage via $A(s_t, a_t) = r_{t+1} + \gamma V(s_{t+1}) - V(s_t)$. The policy is then updated via a clipped objective function which 
allows for efficient updating of the policy, without taking too large of a step in the wrong direction. 
Another important stochastic actor critic algorithm is Soft Actor Critic (SAC) \citep{haarnoja2018soft}. As an entropy maximization algorithm, this has a different objective function than previous examples with an added term $\mathcal{H}(\pi)$, characterizing the entropy of this policy. When maximizing the Q function we can use the soft Q Bellman error 
and the policy can be done via soft policy updates, both of which rely on the added term of the negative log of the policy distribution to maximize for entropy. Although actor-critic algorithms are more common, uncertainty estimation generally occurs on the critic/Q-function. 


All of the above algorithms were designed for online reinforcement learning (in which the agent primarily learns through its own interactions with the environment). However, another field of reinforcement learning has grown in prominence: offline reinforcement learning. In this case the agent has access to a large static dataset of tuples of observations, actions, and rewards and minimal (or no) access to training in the environment. This has a number of challenges that extend beyond the typical challenges of fitting a dataset \cite{levine2020offline}. Uncertainty estimation has proved to be essential for this subfield of RL and many deep RL algorithms for offline RL have some sort of implicit or explicit uncertainty estimation.

\section{Uncertainty in Deep Reinforcement Learning}

When discussing uncertainty we can often decompose it into two sources: aleatoric and epistemic. Aleatoric uncertainty, originating from the Latin \textit{alea} meaning dice, also called statistical uncertainty, is uncertainty that originates from the stochastic nature of the environment and interactions with the environment. Although this uncertainty can be modelled and evaluated, it cannot be reduced. For example, Chess is a game that has zero aleatoric uncertainty whereas Poker has important aleatoric uncertainty. Epistemic uncertainty, also called model uncertainty or systematic uncertainty, is uncertainty that originates from the current limitations of the training of the neural network. Epistemic uncertainty is reducible. For example, the epistemic uncertainty of a classifier is high early in the training but should decrease with more iterations over more data. It is important to acknowledge that the division of these uncertainties are not absolute but represent helpful context dependent heuristics \citep{hullermeier2021aleatoric}. As an example, consider adding a feature (and embedding the problem into a higher dimension), that results in separable data. This `reduces' the aleatoric uncertainty, but also modifies the problems, highlighting the importance of context. 

\subsection{Aleatoric Uncertainty}

Aleatoric uncertainty stems from the behavior and interactions of the environment the reinforcement learning agents are trained on. Hence, the importance of aleatoric uncertainty varies substantially depending on the application. There are 3 main potential sources of aleatoric uncertainty in reinforcement learning (effectively one for each component of the MDP): stochastic rewards, stochastic observations, and stochastic actions. If the reward function is stochastic, there is irreducible uncertainty regarding the true value. The stochastic observations can stem from incomplete observations or stochastic transition dynamics. Incomplete observation games (also called imperfect information games) are extremely common and represent some of the biggest challenges to RL, such as StarCraft II, Dota 2, etc. No amount of training can make an agent see through the fog of war in StarCraft II, hence why this is an example of irreducible (aleatoric) uncertainty.  
If the $P$ function in the MDP is non-deterministic, then the transition from one state to the next is a source of aleatoric uncertainty. In cases like this, for example Poker, Blackjack, etc., the uncertainty what the next observation will be is still there, but in some cases this transition function (although uncertain) can be calculated. Finally, if the actions are stochastic, there is uncertainty about what the next state will be (since the action is uncertain). The obvious example would be any stochastic policy algorithm (PPO, SAC) in which the action is chosen from a distribution instead of a deterministic point.

While seemingly unwanted, aleatoric uncertainty can be injected into games to improve them as benchmarks. For example, Atari games are deterministic, and thus have zero aleatoric uncertainty (assuming the observation representation is markovian). However, fully deterministic single player games are problematic as benchmarks since the agents can just memorize the correct set of steps to maximize reward (rather than actually `learning' to play the game) which has led to forced insertion of aleatoric uncertainty into the environments via techniques such as sticky actions, in which there is a chance that the same action will be used regardless of the agent's input \citep{machado2018revisiting}. This forced action uncertainty makes for a more difficult (and effective) benchmark. Although all three sources of aleatoric uncertainty can be estimated and potentially quantified they cannot be reduced like epistemic uncertainties. That does not mean that awareness is unimportant. Differentiating between areas of high epistemic uncertainty and high aleatoric uncertainty can be essential in the training of an agent. An area of high uncertainty in an environment may want to be explored, but if the uncertainty stems entirely from aleatoric sources, it will result in ineffective training to continue to visit that area (since the agent is already knowledgeable of that area, there just isn't enough information to make a certain decision). Hence, aleatoric uncertainty awareness is crucial for deep reinforcement learning algorithms, even if it cannot be reduced. 

\subsection{Epistemic Uncertainty}

In deep reinforcement learning, uncertainty is connected with some of the foundational problems. For example, the trade-off of exploration versus exploitation in reinforcement learning, in which an agent must decide whether to explore new policies and potentially get a lower reward (or potentially discover a better policy) or exploit its current policy to get a known reward but have the opportunity cost of missing out on a potentially better path. The challenge of effective exploration is connected to epistemic uncertainty. As a helpful model, let us consider a localized epistemic uncertainty, i.e. rather than epistemic uncertainty over the whole environment consider the epistemic uncertainty with respect to a specific region of the state space. In a deterministic game, the epistemic uncertainty of a subset of the state (or observation) space correlates with the exploration (or lack thereof) of the subset. High epistemic uncertainty means the agent is uncertain about the policy (or value) in that subspace, hence the region is underexplored. This relationship is often implicit in exploration strategies for deep reinforcement learning (and will be discussed more in next section). Even basic benchmarks like Atari required implicit uncertainty awareness to achieve human level performance \cite{badia2020agent57} as basic algorithms (PPO, SAC, DQN, etc.) are insufficient.

Epistemic uncertainty is also connected to some of the limitations mentioned in the reinforcement learning section. If we have a neural network learning to approximate a Q function a common problem is the overestimation of the Q values. This overestimation is a result of the fact that the estimates of the Q values are noisy and $\mathbb E [ \max Q ] \geq \max \mathbb E [Q]$, thus resulting in Q value estimates that can be substantially larger than the true values. This highlights the challenges of reducing epistemic uncertainty, which is not always trivially reducible through more data or training. DQNs can experience drops in performance over long training times due to catastrophic forgetting 
(i.e. forgetting how to perform well early on in the environment because it has been training primarily on observations from late in the environment),
so simply running the training algorithm for longer will not always decrease epistemic uncertainty. This highlights that even in fully deterministic environments, uncertainty can not be trivially wished away and there is a need for uncertainty aware algorithms. 

\begin{table*}[t]
\small
\begin{tabular}{l c c c c c c} 
 \hline
 Paper & Online/Offline & Uncertainty Method & Base Algorithm & Type of Uncertainty  \\ [0.5ex] 
 (Moerland et al. 2017) & Online & MC-Dropout \& Variance Networks & Distributional DQN & Epistemic \& Aleatoric \\
  \cite{kalweit2017uncertainty}  & Online & Bootstrapped Q & Model Based DDPG & Epistemic \\ 
  (Osband et al. 2018) & Online & Bootstrapped Q \& Prior Networks & DQN & Epistemic \\
   \cite{clements2019estimating} & Online  & Bootstrapped Q (from MAP) & QR-DQN & Epistemic \& Aleatoric \\
    \cite{yu2020mopo} & Offline  & Variance of Dynamics Model & SAC  & Epistemic \\
\cite{peer2021ensemble} & Online & Bootstrapped Q & DQN & Epistemic \\
 \cite{an2021uncertainty} & Offline & Bootstrapped Q & SAC  & Epistemic \\
 \cite{wu2021uncertainty} & Offline  & MC-Dropout & Actor-Critic  & Epistemic \\
 \cite{hiraoka2022dropout} & Online & MC-Dropout \& Bootstrapped Q & SAC & Epistemic \\
 \cite{mai2022sample} & Online & Bootstrapped Variance Q Networks & SAC  & Epistemic \& Aleatoric \\

 \cite{lee2022offline} & Offline & Bootstrapped Q and Policy & SAC + CQL  & Epistemic \\
 \cite{bai2022pessimistic} & Offline & Bootstrapped Q & SAC  & Epistemic \\
 (Ghasemipour et al. 2022) & Offline & Bootstrapped Q & Actor-Critic & Epistemic \\
\cite{mavor2022stay} & Online & Variance Networks & PPO & Aleatoric \\
 \hline
\end{tabular}
\caption{A Selection of Works in Uncertainty Aware Deep Reinforcement Learning}
  \label{tab:1}
\end{table*}

\subsection{Exploration and Uncertainty}

Research into epistemic uncertainty for reinforcement learning has figured prominently in exploration research, although it is not always explicitly mentioned. There are many ways to estimate epistemic uncertainty, and a number of works implicitly rely on uncertainty estimation. Although we will focus on explicit uncertainty calculations later in this paper, we highlight a few prominent implicit calculations (or calculations by proxy) here since exploration remains an active and important area of research and understanding these connections can help provide insight into both exploration and uncertainty quantification. One such exploration technique is count based methods. Count based methods have a simple origin, assuming our agent can learn equally well from all examples, if we want to minimize epistemic uncertainty we need to provide training information from all parts of the state space. Therefore, the areas with the highest epistemic uncertainty are the areas with the least training examples. If we simply count and keep track of how often every state has been trained on, we can easily minimize uncertainty (by seeking out the least visited states) and efficiently explore. However, this quickly becomes infeasible for large state spaces.

Contemporary count based methods attempt to approximate this ideal foundation using more tractable methods. In one of the foundational methods of count based exploration, \citep{bellemare2016unifying} presented a method based on density models and pseduo-counts.  Specifically, a density model provides the ratio between the pseudo-count function, $\hat{N}(x)$ and the pseudo-count total $\hat{n}$. These pseduo-counts are estimates of the true count function through the proxy of the density model (since $\hat{N}$ is really the function we care about). This density model $\rho_i(x)$ is a function of a given observation x, that it has encountered $i$ times. This model must satisfy the inequality $\rho_{i+1}(x)  = \frac{\hat{N}_{i+1}(x) + 1}{\hat{n} + 1} \geq \rho_{i}(x)  = \frac{\hat{N}_{i}(x)}{\hat{n}}$. We can then approximate the value of interest $\hat{N}_i(x)$
from this density model alone. This count is then used as an intrinsic reward to direct the model to areas of high uncertainty. Another way to approximate this count function is by mapping high dimensional observations to lower dimensional representations (then using these representations to count with) \cite{tang2017exploration}. 
The principle remains the same: to approximate the uncertainty from the data side, i.e. by counting (or approximately counting) the amount of training done on different areas of the observation space. Note that this is an approximation of the uncertainty, and the quality of this approximation is dependent on the density model's task relevance \citep{osband2019deep}. Another way to approximate this uncertainty uses a neural network to learn to approximate this uncertainty. One such example is Random Network Distillation (RND) that uses a static neural network that is a function of the state and another neural network that is trained to predict the output of the static neural network \cite{burda2018exploration}. This loss is used as an exploration bonus (as the more times the state has been visited, the more opportunities there are to learn and the lower the loss will be) and as a proxy for model uncertainty.

\section{Addressing Uncertainty in Reinforcement Learning}

A overview of the key papers we outline in this section can be found in Table \ref{tab:1}. The base algorithm is the deep reinforcement learning algorithm that the uncertainty quantification is incorporated into. An empirical comparison of the offline algorithms is presented in Table \ref{tab:res}. Behavioral Cloning (BC) 
, Conservative Q Learning (CQL) \cite{kumar2020conservative}, and Twin Delayed Deep Deterministic Policy Gradient Behavioral Cloning (TD3 + BC) \cite{fujimoto2021minimalist} are used as baselines. All results are taked from the paper they originally appeared in. Note that \cite{lee2022offline} is not included as it focuses on the transition from offline to online RL. In \cite{ghasemipour2022so}, the results are only presented in graph form, so values may not be exact. The results are only presented for offline RL due to the ubiquity of the D4RL benchmark. Although there are popular benchmarks in online RL, every paper does not have the exact same environments. The papers represented in this work should provide a large spectrum of techniques and algorithms to help provide insight into the breadth of research in uncertainty aware RL. 

\subsection{Common Techniques}

Now let us consider how we can address the aforementioned sources of uncertainty. First, let us note that by variance networks, we mean neural networks with heads $f$ and $\phi$, the latter of which directly outputs an estimate of the variance. Two of the most common and important techniques for these approaches are bootstrapping and Monte Carlo (MC) dropout. The Bootstrapped DQN was introduced by \cite{osband2016deep} as a method for efficient exploration. This method is a modification of the traditional DQN neural network architecture that uses a shared torso with $K, K \in \mathbb N$ heads (in the original paper $K = 10$). 
By computing the variance of the heads predictions, we have an estimation for the posterior. Since the heads are picked for training uniformly, they will ideally overlap only when they have all arrived at the optimal Q function. This method provides an estimation of epistemic uncertainty and will form the basis of many more advanced techniques.

Dropout 
is a method to reduce overfitting in neural networks by stochastically dropping (setting the weight to 0) neurons. This dropout is traditionally done during training and removed during inference. It is common knowledge that using dropout for neural networks in reinforcement learning is a bad idea (due to the non-stationary targets and the lack of concerns with direct overfitting). However, dropout can be used as a method to approximate model (epistemic) uncertainty. MC Dropout involes keeping the dropout layers during inference and sampling repeatedly, through which one can attain an estimate of the uncertainty by looking at the variance of these predictions \citep{gal2016dropout}. 
This variance serves to estimate the epistemic uncertainty. 
Both of these methods are conceptually straightforward and relatively easy to implement, making them a strong basis for many approaches in deep reinforcement learning.

\begin{table*}[t]
\centering
\small
\begin{tabular}{l c c c c c c c c} 
 \hline
 Environment & BC & CQL & TD3 + BC & \citeauthor{yu2020mopo} & \citeauthor{an2021uncertainty} & \citeauthor{wu2021uncertainty} & \citeauthor{bai2022pessimistic} & Ghasemipour et al.*   \\ [0.5ex] 
Random HalfCheetah & $2.1$ & $\mathbf{35.4}$  & $10.2$ & $\mathbf{35.4}$ & $28.4$ & $14.5$ & $13.1$ &   \\
Random Hopper & $9.8$ & $10.8$  & $11.0$ & $11.7$ & $25.3$ & $22.4$ & $\mathbf{31.6}$      \\
Random Walker2D & $1.6$ & $7.0$  & $1.4$ & $13.6$ & $\mathbf{16.6}$ & $15.5$ & $8.8$          \\
Medium HalfCheetah & $36.1$ & $44.4$  & $42.8$ & $42.3$ & $65.9$ & $46.5$ & $58.2$ & $\mathbf{70}$ \\
Medium Hopper & $29.0$ & $58.0$  & $99.5$ & $28.0$ & $\mathbf{101.6}$ & $88.9$ & $81.6$ & $75$ \\
Medium Walker2D & $6.6$ & $79.2$  & $79.7$ & $17.8$ & $\mathbf{92.5}$ & $57.5$ & $90.3$ & $80$ \\
Expert HalfCheetah & $107.0$  & $104.8$  & $105.7$ & & $106.8$ & $\mathbf{128.6}$ & $96.2$ & $98$ \\
Expert Hopper & $109.0$ & $109.9$  & $112.2$ & & $110.1$ & $\mathbf{135.0}$ & $110.4$ & $108$ \\
Expert Walker2D & $125.7$ & $\mathbf{153.9}$  & $105.7$ & & $115.1$ & $121.1$ & $109.8$ & $112$ \\
Medium Expert HalfCheetah & $35.8$ & $62.4$  & $97.9$ & $63.3$ & $106.3$ & $\mathbf{127.4}$ & $93.6$ & $96$ \\
Medium Expert Hopper & $111.9$ & $110.0$  & $112.2$ & $23.7$ & $110.7$ & $\mathbf{134.7}$ & $111.2$ & $110$ \\
Medium Expert Walker2D & $6.4$ & $98.7$  & $101.1$ & $44.6$ & $\mathbf{114.7}$ & $99.7$ &  $109.8$ & $112$ \\
 \hline
\end{tabular}
\caption{Empirical Comparison on subset of D4RL of Uncertainty Aware Offline Methods}
  \label{tab:res}
\end{table*}

\subsection{Bootstrapping}

Here we highlight a set of papers that primarily rely on bootstrapping to estimate the uncertainty. We provide a brief overview of what the work did and show how it fits into the broader narrative. 
\cite{kalweit2017uncertainty} replaces the vanilla deep Q networks with bootstrapped DQNs in model based DDPG. They showed that this approach can be up to 15 times as efficient and achieve hundreds of times larger rewards on continuous control tasks such as Reacher. This is representative of a common trend (although sometimes presented with other modifications), that simple incorporation of existing uncertainty aware techniques can yield empirical gains.

Although the majority of approaches are focused on estimating epistemic uncertainty, it can be important to decouple the sources of uncertainty. This idea of decoupled uncertainty awareness is explored by \citep{clements2019estimating}. In this work, their estimates rely on approximate maximum \textit{a posteriori} (MAP) sampling \citep{pearce2020uncertainty} to generate two sets of neural network parameters $\theta_a, \theta_b$. The epistemic uncertainty can then be evaluated via 
the mean squared difference in the predictions of the sampled parameters. Hence, when the variance in the sampled parameters decreases (as the neural network converges), this approximation of epistemic uncertainty will decrease. Although this is different than other bootstrapped approaches, it is a variant of the same approach. The MAP serves effectively to generate better versions of the heads from Bootstrapped DQN, hence requiring only two of them (as opposed to 10 or more). The aleatoric uncertainty is estimated via the covariance of these predictions.
These parameters are then used in action selection (and action selection only), and this uncertainty is not evaluated directly in the loss functions/training of the neural network. They are able to show substantial improvements on the whole of the MinAtar environments (smaller versions of the Atari benchmark). This shows two key results: that uncertainty awareness can yield benefits even when used in action selection alone and the potential of decoupling uncertainties even in environments which do not seem to present much uncertainty (e.g. deterministic Atari games). 

\cite{peer2021ensemble} builds upon both BDQN and DDQN. They present a DQN variant which uses $K$ independent networks. Whereas BDQN only uses separate heads for each prediction, this work uses entirely separate networks. While DDQN uses separate networks only for target prediction, this work uses them for action selection as well. By simply using $K = 5$ networks, they are able to achieve comparable performance to Rainbow \cite{hessel2018rainbow} on Atari. The importance of this result is its conceptual simplicity. Such a simple DQN expansion is able to compete with Rainbow (i.e. all previous combined DQN modifications).

\cite{an2021uncertainty} added an ensemble of $N$ Q function on SAC (which usually has 2 Q functions), to enable a bootstrapped estimate of epistemic uncertainty. The agent's understanding of epistemic uncertainty enabled it to perform well on Out of Distribution (OOD) data (which is important for offline RL). OOD data is data that is collected under different conditions than the training data. This simple modification of adding Q functions (although it sometimes required up to 500), shows how a conceptually simple uncertainty aware modification can increase performance. Expanding upon this, Ensemble-Diversified Actor Critic (EDAC) was also introduced, which is extremely similar to $N$-SAC but adds a term to the soft Q update to increase the variance of the Q functions when encountering OOD data. Conceptually, this is simply increasing the accuracy of the epistemic uncertainty approximation (since the model should have high epistemic uncertainty when it encounters unfamiliar data/situations). They were able to show state of the art performance at the time of their writing (and remains competitive even with current works) and improve upon a heavily model based field. 

\cite{lee2022offline} used bootstrapped uncertainty estimates of the both the actor and critic and combined them with Conservative Q Learning \citep{kumar2020conservative} to achieve improvements on the transition of offline to online continuous control tasks. Although the most common choice of combination (i.e. the algorithm the uncertainty estimation is incorporated into) is SAC, this highlights how different choices for this algorithm can impact the efficacy of uncertainty estimation.

\cite{bai2022pessimistic} uses the standard deviation of bootstrapped prior deep Q networks as an uncertainty quantification. This uncertainty is then subtracted from the predicted Q value of the next state (essentially penalizing high uncertainty states). This is done for both in distribution (ID) data and OOD. The ID data is the data provided in the dataset and they generate OOD data by feeding a state into the actor (which will predict something different than the dataset, hence OOD) and estimate both the ID Q errors and OOD Q errors to minimize their sum. This uncertainty quantification is then combined with SAC. This highlights the versatility of uncertainty estimation and how figuring out the optimal way to incorporate and learn from this uncertainty remains very much an open problem. 

\cite{ghasemipour2022so} further develops a simple but important aspect of pessimistic updates: target independence. Pessimistic updates, which are common in offline RL and exploration techniques, usually relies on subtracting the epistemic uncertainty estimation via variance of Q values from the predict Q values. Shared targets use pessimistic predictions for the target values in the Q learning updates, whereas independent targets do not. These independent targets have favorable theoretical implications and are shown to have empirical benefits for offline RL. Strong theoretical understanding is important to advancing and developing further uncertainty aware RL algorithms.

\subsection{Monte Carlo Dropout}

Here we highlight papers that primarily rely on MC-Dropout to estimate the uncertainty.
In \cite{moerland2017efficient}, they present a method for evaluating the epistemic and aleatoric uncertainties simultaneously and use these methods as exploration strategies. They estimate epistemic uncertainty using MC-dropout and predict the aleatoric uncertainty over the return distribution $p(Z|s,a)$ by having the neural network output $\mu = Q(s,a)$ and $\sigma$, which results in the modified Bellman error with an added term to penalize substantial changes in aleatoric uncertainty and subtly encourages lower aleatoric uncertainty. 
This is able to show some modest improvements on a variety of simple gym benchmarks. Although the empirical results are modest, this work presents important ideas that other papers have built upon.

\cite{wu2021uncertainty} presents a method of uncertainty aware offline reinforcement learning in which MC-dropout is used to estimate epistemic uncertainty. They use MC-droupout to reduce the effect OOD backups have by scaling each the actor and critic errors by the same amount, $\frac{\beta}{Var(Q(s, a))}$, i.e. by the inverse variance of the Q functions (the variance calculated across the MC estimates). This means that high variance estimates (high epistemic uncertainty) will contribute less to the update process. This also achieves improvements on offline continuous control benchmarks, but generally only outperforms SAC-$N$ and EDAC on the expert replay examples (i.e. when the offline dataset contains transitions of a expert completing the task, as opposed to medium skill level or random). Along with \cite{an2021uncertainty}, this work remains extremely competitive with current offline RL works. Since MC Dropout is a less common approach, this work shows that it is still able to be as effective as bootstrapped methods.

\subsection{Combinations and Other Approaches}

Here, we present a collection of methods that do not solid rely on a single method or use a method not previously discussed. \citep{osband2018randomized} highlights some of the flaws in the previously mentioned methods and introduces the concept of randomized priors as an add-on to bootstrapped DQN. First, they show that MC-dropout (and the distribution it generates) is not always a good estimate of the posterior. Additionally, they highlight the problem with count based estimates mentioned previously. Specifically, that they can be a very poor proxy for uncertainty for cases in which state density does not correlate with the true uncertainty. To overcome these challenges they suggest simply adding a random prior to bootstrapped DQN. This operates in a unique way from the previously discussed techniques since it doesn't focus on estimating the posterior. What this means is that all previous methods relied on estimating uncertainty based on the data, whereas the prior is independent of the data. This is an important aspect of the Bayesian approaches that these methods are often claiming to approximate. Practically, this is simply a matter of creating $K$ prior neural networks (with static weights) that are then incorporated into the bootstrap heads so the final output is $Q(s, a) = f_k(\theta, s, a) + p_k(\phi, s)$. They show that this results in improved exploration by achieving state of the art (at the time) performance on the hard exploration Atari game Montezuma's Revenge.

In one of the foundational works of offline RL, \cite{yu2020mopo} bootstrapped a prediction of the next reward using an ensemble of neural networks and penalized the reward that is learned by the policy proportionally to the maximum standard deviation of the predictions in this ensemble. This is different than the previous approaches to bootstrapping since it does not rely on Q function estimates. This approach achieved substantial improvements across the board on a variety offline continuous control tasks. 

In \cite{hiraoka2022dropout}, the ideas of dropout and bootstrapped Q values are combined to modified the Q network architecture to have several heads with dropout layers
. By combining the uncertainty estimation techniques, this results in a higher quality approximation of the epistemic uncertainty. These updated Q networks are then used as critics with SAC to achieve better performance substantially faster (around 100,000 frames as opposed to the usual 1-3 million) on the MuJoCo continuous control benchmark.

\cite{mavor2022stay} present a unique approach to aleatoric uncertainty, by using a neural network with two heads that directly output the predicted mean and variance of the next state. This network variance is a learned estimation of the aleatoric uncertainty and is updated using the Maximum Likelihood Estimation (MLE) loss function, similar to the work done in \cite{kendall2017uncertainties}. They are able to show substantial improvements in \say{noisy TV} environments (i.e. environments with large aleatoric uncertainty traps). This represents a far different approach to uncertainty estimation and highlights the relationship between uncertainty estimation and exploration.

Building upon many of the above techniques and approaches from supervised learning, \cite{mai2022sample} present a method for \say{inverse-variance} reinforcement learning with decoupled uncertainty estimates. Specifically, they modify the value function loss over a minibatch to be $\mathcal{L} = \mathcal{L}_{BIV} + \lambda \mathcal{L}_{LA}$. Here, $\mathcal{L}_{BIV}$ is the Batch Inverse Variance (BIV) 
which weights the loss function 
inversely proportional to the variance of the noise of the label. This label noise is representative of the aleatoric uncertainty (which is estimated through the variance networks). The $\mathcal{L}_{LA}$ is the loss function of variance networks, i.e. the negative log likelihood of the neural network outputs $\mu$ and $\sigma$ (which can estimate epistemic uncertainty). By using an ensemble of variance networks, they can bootstrap an estimate of the uncertainty for the BIV loss function. They apply this uncertainty awareness to the critics of SAC and show substantial improvements on a variety of continuous control tasks.

\section{Discussion}

First, it is worth noting that most approaches to uncertainty estimation in deep RL focus on the critic (which is also the policy in the case of DQN variants). This may seem counter-intuitive, seeing as it is the policy which actually makes decisions and operates in the uncertain world. However, there are likely two important reasons why uncertainty estimation for the critic prioritized. For one, during training the critic is more directly a influenced by the aleatoric uncertainty since only the critic loss is a direct function of the state, action, and reward. Second, the actor is downstream of the critic in uncertainty propagation. Since all actor-critic algorithms updates take some form of $\nabla J = \mathbb E [\nabla log (\pi) \delta]$, with $\delta$ being some function dependent on the critic, any uncertainty that exists in the critic is passed down to the actor. This does not preclude uncertainty considerations for the actor (and this is likely the directions of many future works in uncertainty estimating RL), but sufficient uncertainty estimation in the critic is important.

Second, a brief postulate as to why MC Dropout is less common than using network heads to bootstrap. MC dropout requires a full network pass for each repetition/sample used to generate the distribution of predictions, whereas bootstrapping from heads only requires multiple forward passes of a subset of layers (i.e. the heads). This can result in a non-negligible compute difference. However, the relative quality of the estimates from bootstrapped heads and full MC Dropout remains an important but under explored area.

Next, let us note some trends that Table \ref{tab:1} and the explanations highlight. First, uncertainty estimation can be incorporated into most (if not all) existing reinforcement learning algorithms, and doesn't require a ground-up re-evaluation. Additionally, it shows how some implementationally simple techniques, can be used to result in large benefits. These techniques are often built on the back of extensive work done in the supervised learning community.  These papers show that uncertainty awareness can help improve a variety of algorithms across types and classes, on different benchmarks with different levels of aleatoric uncertainty.

Lastly, it is important to highlight some of the important takeaways for applications of reinforcement learning, in addition to some potential future work. 
Even though many games involve a substantial amount of uncertainty, quantification often drops by the wayside since basic algorithms perform moderately well. 
Incorporating uncertainty into algorithms has shown empirical improvements on a variety of tasks, even those that are largely deterministic. 
Given the potential benefits (as we have shown throughout) and the ease of implementation, it is our goal to get a larger community to consider the incorporation of some of these methods into their practices. 
There is also room for improvement in the techniques presented in this work. Further empirical insight is needed into the relative performance of different uncertainty quantification methods, since these techniques are often combined with other changes it is hard to provide a meaningful comparison between advancements. It is also important to quantify how good the uncertainty estimates are. Seeing how close the estimates are to true values would enable better comparisons of methods. These also speak to the need to develop environments that enable the testing and evaluation of uncertainty estimation. More work formalizing and experimenting with the relationship between epistemic uncertainty and common exploration strategies could also be beneficial.

\section{Conclusion}

Although deep reinforcement learning has become the ubiquitous method to solve a variety of control tasks and games, it still suffers from problems such as brittleness and data inefficiency. Understanding and incorporating uncertainty into deep RL agents is critical to their success in games and real world situations and their robustness to new environments. We outlined the basics of contemporary deep RL algorithms and highlighted the relationship of RL with uncertainty. Outlining the roles of aleatoric and epistemic uncertainty, we showed how these connect with the basic ideas in the field. We presented a variety of techniques for working with and incorporating awareness of these uncertainties into reinforcement learning algorithms. We show that these techniques are able to improve performance across domains, benchmarks, models, and types of reinforcement learning. 

\bibliography{aaai22}

\end{document}